\def\year{2018}\relax
\documentclass[letterpaper]{article} 
\usepackage{aaai18}  
\usepackage{times}  
\usepackage{helvet}  
\usepackage{courier}  
\usepackage{url}  
\usepackage{graphicx}  
\usepackage{amsmath}
\usepackage{verbatim}

\usepackage{ifthen}
\usepackage{color}

\begin{document}
\title{Self-view Grounding Given a Narrated 360$^{\circ}$ Video
}
\author{Shih-Han Chou$^{\dagger}$,\ Yi-Chun Chen$^{\dagger}$,\ Kuo-Hao Zeng$^{\dagger}$,\ Hou-Ning Hu$^{\dagger}$,\ Jianlong Fu$^{\ddagger}$,\ Min Sun$^{\dagger}$\\
$^{\dagger}$Department of Electrical Engineering, National Tsing Hua University\\
$^{\ddagger}$Microsoft Research, Beijing, China\\
{\tt\small \{happy810705, yichun8447\}@gmail.com, khzeng@cs.stanford.edu}\\
{\tt\small \{eborboihuc@gapp, sunmin@ee\}.nthu.edu.tw, jianf@microsoft.com}}

\maketitle
\begin{abstract}
Narrated 360$^{\circ}$ videos are typically provided in many touring scenarios to mimic real-world experience. However, previous work has shown that smart assistance (i.e., providing visual guidance) can significantly help users to follow the Normal Field of View (NFoV) corresponding to the narrative.
In this project, we aim at automatically grounding the NFoVs of a 360$^{\circ}$ video given subtitles of the narrative (referred to as ``NFoV-grounding"). We propose a novel Visual Grounding Model (VGM) to implicitly and efficiently predict the NFoVs given the video content and subtitles. Specifically, at each frame, we efficiently encode the panorama into feature map of candidate NFoVs using a Convolutional Neural Network (CNN) and the subtitles to the same hidden space using an RNN with Gated Recurrent Units (GRU). Then, we apply soft-attention on candidate NFoVs to trigger sentence decoder aiming to minimize the reconstruct loss between the generated and given sentence. Finally, we obtain the NFoV as the candidate NFoV with the maximum attention without any human supervision.
To train VGM more robustly, we also generate a reverse sentence conditioning on one minus the soft-attention such that the attention focuses on candidate NFoVs less relevant to the given sentence. The negative log reconstruction loss of the reverse sentence (referred to as ``irrelevant loss") is jointly minimized to encourage the reverse sentence to be different from the given sentence. 
To evaluate our method, we collect the first narrated 360$^{\circ}$ videos dataset and achieve state-of-the-art NFoV-grounding performance.
\end{abstract}

\section{Introduction}
\begin{figure}[t]
  \centering
    \includegraphics[width=0.45\textwidth]{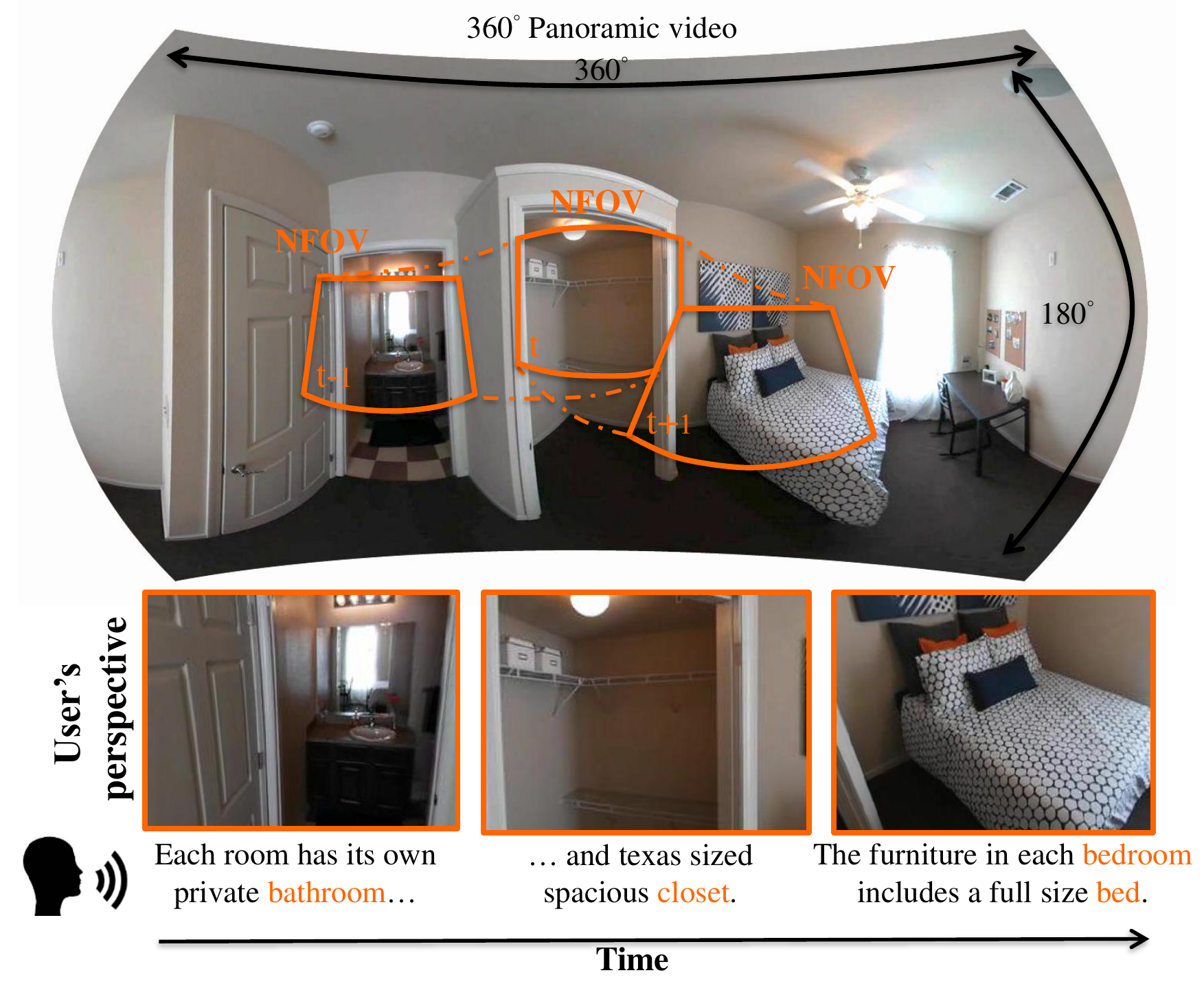}
    \caption{\small \textbf{Illustration of NFoV-grounding}. In a 360$^{\circ}$ video (top-panel), a video player displays a predefined Normal Field of View (NFoV) (bottom-panel). Our VGM can automatically ground narrative into a corresponding NFoV (orange boxes) at each frame.}\label{Fig.1}
\end{figure}


Thanks to the availability of consumer-level 360$^{\circ}$ video cameras, many 360$^{\circ}$ videos are shared on websites like YouTube and Facebook. Among these videos, a subset of them is narrated with natural language phrases describing the video content. For instance, many videos consist of the guided tour of real estates (e.g., houses and apartments) or tourist locations. Intuitively, a 360$^{\circ}$ touring video provides viewers the freedom to follow the narrative and select Normal Field of Views (NFoVs) provided by typical video players (see Fig.~\ref{Fig.1}).
However, a study in Lin et al.~\shortcite{lin2017tell} suggests that visual guidance (i.e., assistance through visual indicator) is preferred to assist viewers to follow the NFoV described by the narrative. 
In order to provide the visual guidance, the NFoV described by the narrative needs to be inferred. We define this new task as ``NFoV-grounding''. 

The task of NFoV-grounding is related to but different from grounding language in normal images in two main ways. 
First of all, the panoramic image in 360$^{\circ}$ video has a large field of view with high resolution. Hence, the regions described by narrative often are relatively small compared to the panoramic image. As a result, grounding language in panoramic image is harder and computationally expensive. Secondly, rather than grounding to an object, our task is grounding to an NFoV, which could correspond to objects and/or scene regions. In this case, existing object proposal methods cannot be leveraged. To the best of our knowledge, NFoV-grounding is a unique task which hasn't been tackled.

We propose a novel Visual Grounding Model (VGM) to implicitly and efficiently predict the NFoVs given the video content and subtitles. Specifically, at each frame, we encode the panorama into feature map using a Convolutional Neural Network (CNN) and embed candidate NFoVs onto the feature map to efficiently process NFoVs in parallel.
On the other hand, the subtitle is encoded to the same hidden space using an RNN with Gated Recurrent Units (GRU). Then, we apply soft-attention on candidate NFoVs to trigger a sentence decoder aiming to minimize the reconstruction loss between the generated and given sentence (referred to as relevant loss). At the end, we obtain the NFoV as the candidate NFoV with the maximum attention without any human supervision.
We emphasize that the training does not require the knowledge of ground truth NFoV. Similar to Rohrbach et al.~\shortcite{rohrbach2016grounding}, our model relies on sentence reconstruction to implicitly infer the NFoV.
In order to further address the challenges in NFoV-grounding, we propose the following techniques to train VGM more robustly.
First of all, we generate a ``reverse sentence" conditioning on one minus the soft-attention such that the attention focuses on candidate NFoVs less relevant to the given sentence. The negative log reconstruction loss of the reverse sentence (referred to as ``irrelevant loss") is jointly minimized to encourage reverse sentences to be different from the given sentences. Secondly, we augment the panoramic images dataset by exploiting rotation invariant property to randomly shift the viewing angles.

To evaluate our method, we collect the first narrated 360$^{\circ}$ video dataset consisting of both indoor and outdoor tourist guides. We also redefine recall and precision on NFoV-grounding as evaluation metrics. Finally, our model achieves state-of-the-art NFoV-grounding performance and can be run at 0.38 fps on $720 \times 1280$ panoramic image.

Our main contributions can be summarized as follows:
\begin{itemize}
\item We define a new "NFoV-grounding" task which is essential to automatic assist watching 360$^{\circ}$ videos.
\item We propose a novel Visual Grounding Model (VGM) to implicitly and efficiently infer NFoV.
\item We introduce a novel irrelevant loss and a 360$^{\circ}$ data augmentation technique to robustly train our model.
\item We collect the first narrated 360$^{\circ}$ video dataset and achieve the best performance.
\end{itemize}
\section{Related Work}

We review related works in virtual cinematography, 360$^{\circ}$ vision and grounding natural language in images and videos.

\subsection{Virtual Cinematography}
A listing of virtual cinematography research \cite{christianson1996declarative,elson2007lightweight,mindek2015automatized} focused on controlling a camera view in virtual/gaming environments and did not take the problem of perception difficulties into consideration. 
Some works \cite{sun2005region,chen2015mimicking,chen2016learning} relaxed such perception assumption. They manipulated virtual cameras within a static video with wide field-of-view of a teleconference, a basketball court, or a classroom, where objects of interest could be easily extracted.

\subsection{360$^{\circ}$ Vision}
Recently, the perception/experience of 360$^{\circ}$ vision is gained numerous interest. 
Assens et al.~\cite{Assens2017SaltiNet} focused on scan-paths prediction on 360$^{\circ}$ images. The network predicts saliency volumes, which are stacks of saliency maps, then the scan-paths were sampled from the volume conditioned on number, location, and duration of fixations.
Lin et al.~\cite{lin2017tell} concluded that the use of Focus Assistance for 360$^{\circ}$ videos will help viewers to focus on the intended targets in videos.
Su et al.~\cite{su2016activity} referred a problem of viewing NFoV in 360$^{\circ}$ videos as Pano2Vid and proposed an offline method handling unedited 360$^{\circ}$ videos downloaded from YouTube. They further~\cite{su2017videography} improved the offline method in threefold. First, they proposed a coarse-to-fine technique to reduce computational costs. Second, they gave diverse output trajectories. Last, they introduced a degree of freedom by zooming NFoV.
In contrast, Hu and Lin et al.~\cite{hu2017deep360pilot} proposed an online human-like agent piloting through 360$^{\circ}$ videos. They argued that a human-like online agent is necessary in order to provide more effective video-watching supports for streaming videos and other human-in-the-loop applications, such as foveated rendering \cite{patney2016perceptually}.
On the other hand, Lai et al.~\cite{Lai360arXiv17} provided an offline editing tool on 360$^{\circ}$ videos, which equips with visual saliency and semantic scene label, helping end users generate a stabilized NFoV video hyperlapse while our method focuses on semantic automatic visual NFoV-grounding.

\subsection{Grounding natural language in images and videos}
The interaction between natural language and vision has been extensively studied over the past years \cite{zeng2016generation,zeng2017leveraging}. For grounding language in images, \cite{johnson2015image} used a Conditional Random Field model to ground a scene graph query of images in order to retrieve semantically related images. 
\cite{karpathy2014deep} reasoned dependency tree relations on images using Multiple Instance Learning and a ranking objective. 
\cite{karpathy2015deep} replaced the dependency tree with a multi-modal Recurrent Neural Network and simply used maximum value instead of ranking objective.
\cite{wang2016learning} proposed a structure-preserving image-sentence embedding method for retrieval problem and also applied it to phrase localization. 
\cite{mao2016generation} and the Spatial Context Recurrent ConvNet (SCRC) \cite{hu2016natural} used a recurrent caption generation network to localize an object with the highest probability by evaluating the given phrase on the set of proposal boxes. 
\cite{rohrbach2016grounding} proposed an attention localization mechanism with an extra text reconstruction task.
\cite{Rong_2017_CVPR} proposed models for both scene text localization and retrieving candidate text regions by jointly scoring and ranking the text outputs.
\cite{Zhang_2017_CVPR} associated image regions with text queries by a discriminative bimodal neural network, with extensive use of negative samples.
\cite{Xiao_2017_CVPR} localized textual phrases in a weakly-supervised setting by learning pixel-level spatial attention masks as phrases localization.
Several representative works on spatial-temporal language grounding are \cite{lin2014visual}, \cite{yu2013grounded}, and 
\cite{Li_2017_CVPR} aimed at tracking objects of interest from a sequence of natural language specification while ours focus on visual NFoV-grounding where might deal with indoor or outdoor scenery and multiple challenging NFoVs of interest.

\section{Method}
\begin{figure*}[!t]
\centering
  \includegraphics[width=0.67\textwidth]{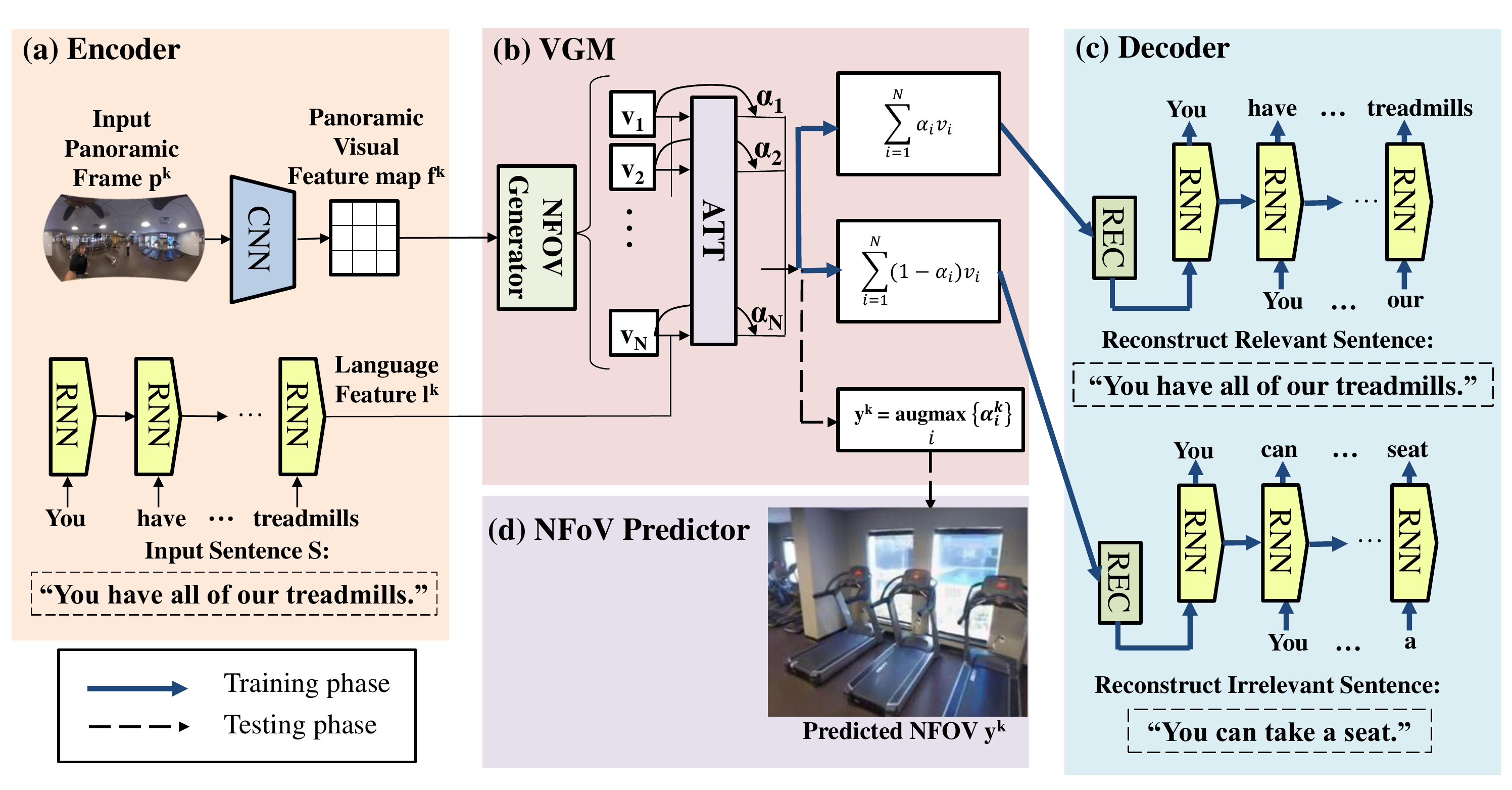}
  \caption{\small Illustration of our method. During unsupervised training, Each 360 video frame pass through the panel (a) panoramic/subtitle encoder, in which given panoramic frames are encoded to visual feature while  corresponding subtitles are encoded to language feature by CNN and RNN respectively, (b) our proposed VGM, in which we propose NFoV candidates and apply the soft-attention mechanism on NFoV candidates with encoded language feature, and (c) language decoder, in which we reconstruct subtitles according to attended feature derived from VGM. In the testing phase, instead of language decoder, (d) NFoV predictor is used. }
  \label{fig_model}
\end{figure*}

Our goal is to build a system that can ground subtitles (i.e., natural language phrases) into corresponding views in each 360$^{\circ}$ video. We define this task as the Normal Field of View (NFoV)-grounding problem since typical 360$^{\circ}$ video players display a predefined NFoV (see Fig. \ref{Fig.1}). We formally define the notation and task below.

\noindent\textbf{Notation.}
We define $\mathbf{p}$ as a sequence of panoramic frames where $\mathbf{p}=\{p^1, p^2, ..., p^k\}_{k=1}^K$ and $K$ is the video length. $\mathbf{f}=\{f^1, f^2, ..., f^k\}$ is defined as the encoded panoramic visual feature map. 
$\mathbf{v}^k=\{v_1^k, v_2^k, ..., v_i^k\}$ is defined as the encoded NFoV candidates' visual feature. $\mathbf{S}$ is the subtitles where $\mathbf{S}=\{\mathbf{s}^1, \mathbf{s}^2, ..., \mathbf{s}^k\}$, $\mathbf{s} = \{s_1, s_2, ..., s_m\}$, and $m$ is the number of words in a subtitle. $\mathbf{L}=\{\mathbf{l}^1, \mathbf{l}^2, ..., \mathbf{l}^k\}$ is defined as the encoded language features, where $\mathbf{l} = \{l_1, l_2, ..., l_m\}$. $\boldsymbol{\alpha}$ is defined as the soft-attention weights where $\boldsymbol{\alpha}^k=\{\alpha_1^k, \alpha_2^k, ..., \alpha_i^k\}$. $v^k_{att}$ is the attended NFoV's visual feature and $\hat{v}^k_{att}$ is the reverse attended NFoV's visual feature.
$v^k_{rec}$ is the reconstructed NFoV's visual feature and $\hat{v}^k_{rec}$ is the reverse reconstructed NFoV's visual feature. 
$\mathbf{y}$ is the predicted NFoVs.
A panoramic frame usually has a corresponding subtitle describing several objects/regions and the relation between them.

\noindent\textbf{Task: NFoV-grounding.}
Given panoramic frames and subtitles,
our work is to ground the NFoV viewpoint based on the subtitles in panoramic frames to provide visual guidance:
\begin{equation}
\mathbf{y}=O(\mathbf{p},\mathbf{S}),
\end{equation}
where $O$ denotes a model inputting panoramic frames $\mathbf{p}$ and subtitles $\mathbf{S}$ and predicting the corresponding NFoVs $\mathbf{y}$. 

\subsection{Model Overview}

We now present an overview of our model. Our model consists of three main components. The first part encodes each panoramic frame and its corresponding subtitle into a hidden space with the same dimension. For visual encoding, every panoramic frame $p^k$ is encoded to a panoramic visual feature map $f^k$ by a convolutional neural network (CNN); for language encoding, the corresponding subtitle $\mathbf{s}^k$ is encoded to language feature $l^k$ by a recurrent neural network (RNN). After encoding, we have the panoramic visual feature map $f^k$ and the subtitle representation $l^k$ (see Fig.~\ref{fig_model}(a)).

The second part is the proposed Visual Grounding Model (VGM) (see Fig.~\ref{fig_model}(b)). Similar to \cite{su2016activity}, we define several NFoV candidates centering at longitudes $\phi \in \boldsymbol{\phi} = \{0, 30, 60, ..., 330\}$ and latitudes $\theta \in \boldsymbol{\theta} = \{0, \pm{15}, \pm{30}\}$.
Then, we propose to encode NFoV visual feature candidate $\mathbf{v}^k$ from panoramic visual feature map $f^k$.
Note that each NFoV corresponds to a rectangle region on an image with a perspective projection, but a distorted region on the panoramic image with equirectangular projection. Hence, we propose to embed an NFoV generator proposed in \cite{su2016activity} into representation space to efficiently encode the panoramic visual feature map $f^k$ into NFoV candidates' visual feature $\mathbf{v}^k$:
\begin{equation}
\label{NFoV_embed}
\mathbf{v}^k = G_{NFoV}(f^k).
\end{equation}
Once we have NFoV candidates' visual feature $\mathbf{v}^k$, the VGM applies the soft-attention mechanism to the encoded NFoV candidates' $\mathbf{v}^k$ guided by the encoded subtitle language features $\mathbf{l}^k$ to obtain the attended weights $\boldsymbol{\alpha}^k$.

The third part is to reconstruct the subtitles. At each frame $k$, we reconstruct the corresponding subtitles by inputting $\boldsymbol{\alpha}^k$ to language decoder during the training phase. Using the attended feature $\boldsymbol{\alpha}^k$ derived from the VGM, we further reconstruct the subtitle. After the reconstruction, we compute the similarity between the original input subtitles and the reconstructed ones. In this case, our goal is to maximize the similarity between them so that we can learn a model specializing in NFoV grounding without direct supervision. On the other hand, during testing phase, we acquire the selected NFoV among $\mathbf{v}^k$ by attention scores during testing phase.

Next, we describe each component in details.

\subsection{Panoramic/Subtitle Encoder}
The goal of this part is to encode the panoramic frame and the subtitle into the same hidden space. Given a 360$^{\circ}$ video panoramic frame and its corresponding subtitles, we use the CNN to encode the 360$^{\circ}$ frame and use the RNN to encode the subtitles. Every panoramic frame is then encoded into a panoramic visual feature map $f$:
\begin{equation}
\label{fCNN}
f^k=CNN(p^k).
\end{equation}
We utilize ResNet-101 \cite{he2016deep} as visual encoder. For subtitle, we first extract word representation by a pre-trained word embedding model\cite{pennington2014glove}. Then we encode the subtitle's representation in turn with an Encoder-RNN ($RNN_{E}$) and obtain the subtitle representation. The language feature $l^k$ is as follow:
\begin{equation}
\label{LRNN}
\mathbf{l}^k=RNN_E(\mathbf{s}^k).
\end{equation}
In practice, we employ Gated Recurrent Units (GRU) \cite{chung2014empirical} as our language encoder.

\subsection{Visual Grounding Model}
This part is the proposed Visual Grounding Model (VGM). The goal of this part is to ground the NFoV in the panorama with the corresponding subtitles. After deriving the encoded panoramic visual feature map $f^k$ and the language feature $\mathbf{l}^k$, we use an NFoV generator to generate the encoded NFoV candidates' visual feature $\mathbf{v}^k$ directly from feature map. The original function of the NFoV generator is to retrieve the pixels in the panoramic frame from a given viewpoint. Proposing NFoV at pixel space and training a visual grounding model like \cite{rohrbach2016grounding} would impose three drawbacks: (1) large memory requirement for the model, (2) considerable time consuming for both training and testing, and (3) making end-to-end training infeasible. As a result, we embed the NFoV generator from pixel space into feature space to mitigate those issues. We propose $60$ spatial glimpses, which are at longitudes $\boldsymbol{\phi}$ and latitudes $\boldsymbol{\theta}$ directly on feature map. Once we have these NFoV candidates' visual feature $\mathbf{v^k}$, we calculate the attention of every NFoV candidate by the corresponding visual feature $v_i^k$ and the subtitle representation $\mathbf{l}^k$. Then, we use the two layers perceptron to compute the attention of each NFoV candidate: 
\begin{equation}
\bar{\alpha}_i^k = ATT(v_i^k, \mathbf{l}^k) = W_a\sigma(W_vv_i^k+W_l\mathbf{l}^k+b_1)+b_a,
\end{equation}
where $\sigma$ is the hyperbolic function, and $W_v$, $W_l$ are the parameters of two fully connected layers. Then we employ softmax to get the normalized attention weights:
\begin{equation}
\alpha_i^k = softmax(\bar{\alpha}_i^k).
\end{equation}

After attaining the attention distribution, we compute attention feature $v^t_{att}$ by calculating the weighted sum of the visual features $v^t$ and the attention weights $\alpha^t$.
\begin{equation}
v_{att}^k = \sum_{i=1}^N \alpha_i^k v_i^k,
\end{equation}
where N is the number of NFoV candidates. 
Furthermore, instead of only reconstructing the subtitles from the most relevant visual feature, we also generate a reverse sentence conditioning on one minus the attention scores such that the attention focuses on irrelevant candidate NFoVs: 
\begin{equation}
\hat{\alpha}_{i}^k=1-\alpha_i^k.
\end{equation}
Then we also calculate the weighted sum to compute the reverse attention feature $\hat{v}_{att}^k=\sum_{i=1}^N \hat{\alpha}_{i}^k v_i^k$.

During testing phase, our model predicts an NFoV $y$ from all NFoV candidates. Because the correct NFoV prediction must contribute the most visual information to reconstruct the corresponding subtitles, our model predicts $y$ by selecting the NFoV having the highest attention score:
\begin{equation}
\label{predict}
y^k=\underset{i}{\mathrm{argmax}}\thinspace\thinspace\{\alpha^k_i\}.
\end{equation}

\begin{figure*}[!t]
  \centering  
\includegraphics[width=1\textwidth]{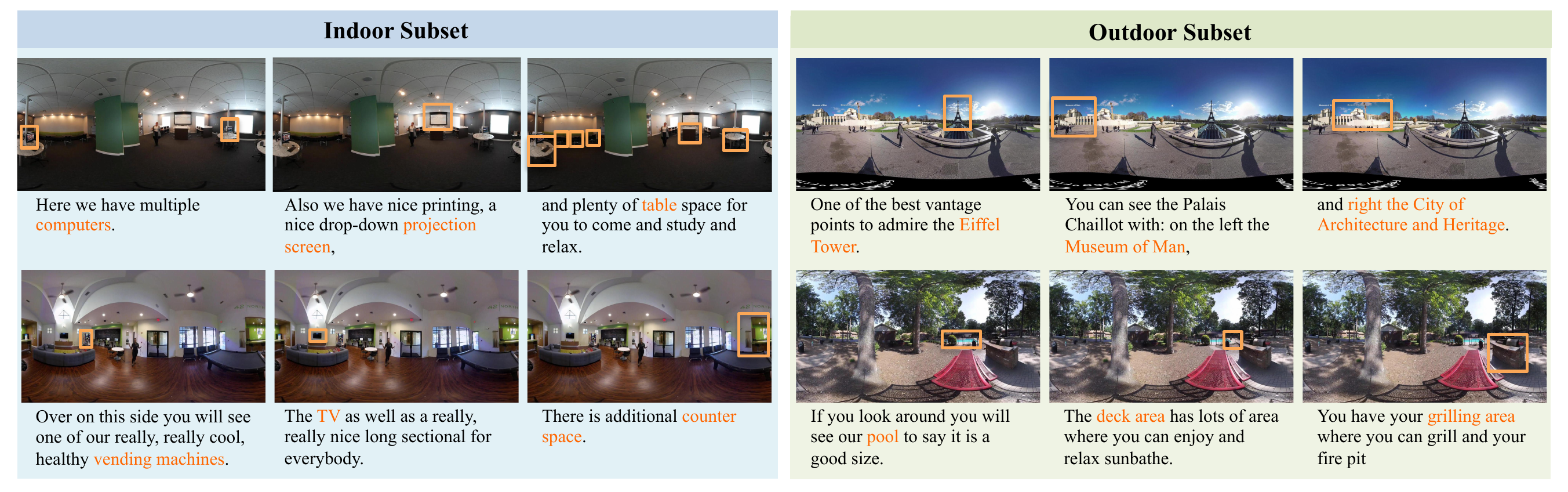}
    \caption{\small Annotation examples. In the left panel shows indoor videos, while outdoor videos are shown in the right panel. Human annotators are asked to annotate the objects or places they would like to see when given the narratives.}
    \label{dataset}
\end{figure*}
\subsection{Reconstruct Subtitles}
As demonstrated in \cite{rohrbach2016grounding}, learning to reconstruct the descriptions of objects within an image shows impressive results on visual grounding task. As a result, we mimic them to employ the reconstruction loss to perform unsupervised learning on NFoV grounding task in $360^{\circ}$ video.

First, we obtain the reconstruct feature $v^k_{rec}$ and the reverse reconstruct feature $\hat{v}_{rec}^k$ after encoding the attention feature by a non-linear encoding layer:
\begin{equation}
\label{eqrec}
v_{rec}^k = \sigma(W_rv_{att}^k+b_r);\hat{v}_{rec}^k = \sigma(W_r\hat{v}_{att}^k+b_r).
\end{equation}
We then employ a Decoder-RNN as our language decoder ($RNN_{D}$). The output dimension of such decoder is set as the size of the dictionary. Thus, our language decoder takes reconstructed feature $v_{rec}^k$ and reverse reconstructed feature $\hat{v}_{rec}^k$ as inputs to generate a distribution over subtitle $s_t$:
\begin{equation}
P(\mathbf{s}_k|v_{rec}^k)=RNN_D(v_{rec}^k); P(\mathbf{s}_k|\hat{v}_{rec}^k)=RNN_D(\hat{v}_{rec}^k),
\label{neg_loss}
\end{equation}
where $P(\mathbf{s}_k|v_{rec}^k)$ and $P(\mathbf{s}_k|\hat{v}_{rec}^k)$ are distributions over the words conditioned on the input reconstructed feature and reverse reconstructed feature, respectively. In practice, we utilize Long Short-Term Memory Units (LSTM) \cite{hochreiter1997long} as our language decoder. Thanks to lots of research on image captioning \cite{vinyals2015show,xu2015show} having demonstrated the effectiveness of LSTM on image captioning task, we first pre-train our language decoder on an image captioning dataset. The pre-trained decoder provides us faster training and better performance on NFoV-grounding task.

Finally, we train our network by maximizing/minimizing the likelihood of the corresponding subtitle $\mathbf{s}_k$ generated via input feature $v_{rec}^k$/$\hat{v}_{rec}^k$ during reconstruction. In this case, the overall loss function can be defined as:
\begin{equation}
\label{loss}
L=\frac{1}{B}\sum^{B}_{b=1}[-{\lambda}P(\mathbf{s}_k|v_{rec}^k)+(1-\lambda)log(P(\mathbf{s}_k|\hat{v}_{rec}^k))],
\end{equation}
where $B$ and $\lambda$ denote batch size and the controlled ratio balancing the effect between relevant and irrelevant visual feature. Because the relevant loss has a lower bound ($zero$), it will converge to the lower bound when we minimize it. However, the irrelevant loss is unbounded ($-infinite$), so when we try to minimize both losses, the irrelevant loss would dominate. Therefore, we mitigate this issue by adding a $log$ function to the irrelevant loss.

\subsection{Training with Data Augmentation on $360^{\circ}$ image}
Since $360^{\circ}$ images are panoramic, we are allowed to arbitrarily augment training data by rotating an entire panoramic image along the longitude. In practice, we online randomly select a value $x \in [0, X]$ where $X$ is the length of a panoramic image. Then, we paste the pixels on the left-hand side of $x$ to the end of the right-hand side. This operation performs the rotating along the longitude centering on $x$. Since we do this operation online, no data is stored in advance. As a result, our augmentation method on $360^{\circ}$ frames provides memory-free data augmentation.

\subsection{Implementation Details}
We train our model only relying on Eq. (\ref{loss}) which does not contain any supervision for NFoV grounding.
We set $\lambda = 0.8$. This is set empirically without heavily tuning. We decrease the frame rate to $1$ to save memory usage and set dictionary dimension as $9956$ according to the number of words appearing in all subtitles. We randomly sample $3$ consecutive frames during training phase (i.e., $k = 3$), but evaluate all frames during testing phase (i.e., $k = \#frames$). Since the maximal length of subtitles is $33$, we set $m = 33$ and give the remaining empty words $Pad$ token if the length of subtitle less than $33$. Besides, we add $Start$ and $End$ tokens to represent the beginning and end of a subtitle, respectively. We use Adam \cite{kingma2014adam} as optimizer with default hyperparameters and $0.001$ learning rate and set batch size $B$ by $4$. We use ResNet-101 pre-training on ImageNet \cite{deng2009imagenet} as our visual encoder and we pre-train our language decoder on MSCOCO dataset \cite{lin2014microsoft}. We implement all of our methods by PyTorch \cite{paszkepytorch}. 
The training phase costs about $17\ min$ and $7\ sec$ per epoch and please refer to the Sec. Modal Efficiency for further model's efficiency analysis.

\section{Dataset}
\begin{table*}[!t]
\caption{\small Our 360 Videos Dataset: we list the statistics information of our dataset. We separate train/val/test set, and they all contain indoor/outdoor set. In the ground truth annotation, we do not label the training set thus the number of it is unknown.}
\centering
\label{my-label}
\resizebox{0.7\textwidth}{!}{
\begin{tabular}{|c|c|c|c|c|c|c|}
\hline
 & \multicolumn{2}{c|}{Train} & \multicolumn{2}{c|}{Validation} & \multicolumn{2}{c|}{Test} \\ \hline
 & Indoor & Outdoor & Indoor & Outdoor & Indoor & Outdoor \\ \hline
Video length (in average) & 16.4 & 13.87 & 18.1 & 15.9 & 23.4 & 22.2 \\ \hline
Video length (maximum) & 44 & 44 & 37 & 35 & 83 & 76 \\ \hline
subtitle length (in average) & 11.6 & 10.64 & 11.4 & 9.67 & 10.5 & 10.9 \\ \hline
subtitle length (maximum) & 33 & 32 & 38 & 29 & 30 & 51 \\ \hline
\#Ground truth annotation (in average) & -- & -- & 2.95 & 1.49 & 3.18 &  1.64\\ \hline
\#Videos & 466 & 216 & 41 & 41 & 56 & 44 \\ \hline
\end{tabular}
}
\end{table*}

In order to evaluate our method, we collect the first narrated 360$^{\circ}$ videos dataset. This dataset consists of touring videos, including scenic spots and housing introduction, and subtitles files, including subtitle text and start and end timecode. Both the videos and the subtitle files are downloaded from Youtube, noted that some of the subtitles are created by Youtube's speech recognition technology.

The videos are separated into two categories: indoor and outdoor, and resized to $720 \times 1280$ first. We extracted a continuous video clip from each video where a scene transition is absent. Subtitle files are also clipped into several files according to the transition time. For training data, we select those whose duration is within 90\% of the range of the duration of all videos, so the max video length is 44 seconds. Also, since the outdoor touring videos are easy to contain some uncommon words or non-English words, our model is hard to learn the word to find the best viewpoint. Hence, we use WordNet as the criterion for outdoor training videos, only sampling the videos without uncommon words (words not in WordNet).  For validation and testing data, both video segments and their annotated ground truth objects are included. We ask human annotators to annotate the objects mentioned in narratives on panoramas chronologically according to the start and end time code in subtitle files. Example panoramas and subtitles are shown in Fig. \ref{dataset}. Finally, we have 563 indoor videos and 301 outdoor videos. We assign 80\% of the videos and subtitles for training and 10\% each for validation and testing. (Available at {\footnotesize \url{http://aliensunmin.github.io/project/360grounding/}})

\section{Experiments}
\begin{table}
\caption{\small Ablation Studies. We evaluate several variants of the proposed model. The w/ denotes with.}
\begin{center}
\resizebox{0.47\textwidth}{!}{
\begin{tabular}{ |c|c|c|c|c| } 
 \hline
 Method / Model &  RL & RL-f & D$^{\dagger}$-RL-f & D$^{\dagger}$-RIL-f \\ \hline\hline
 w/ Relevant Loss & $\surd$ & $\surd$ & $\surd$ & $\surd$ \\ \hline
 w/ Embedded NFoV generator & -- & $\surd$ & $\surd$ & $\surd$ \\ \hline
 w/ Data augmentation & -- & -- & $\surd$ & $\surd$  \\ \hline
 w/ Irrelevant Loss & -- & -- & -- & $\surd$ \\ \hline
\end{tabular}
}
\end{center}
\label{ablation}
\end{table}

Because the style of indoor videos and outdoor videos are different on both vision and subtitles, we first conduct the ablation studies of our proposed method and compare our model with baselines in the beginning. Then, we compare our best model with baselines on the total dataset (i.e., the combination of indoor and outdoor subsets) to demonstrate the robustness of our proposed method. In the end, we manifest the efficiency of our proposed method by measuring the speed of our best model. In the following, we first describe the baseline methods and variants of our method. Then, we define the evaluation metrics. Finally, we show the results and make a brief discussion.

\subsection{Baselines}
\noindent \emph{- RS:} Random Selection. We randomly select the NFoV candidates for fundamental evaluation.

\noindent \emph{- CS:} Centric Selection. We evaluate the dataset by selecting centric NFoV candidate as predicted NFoV. The effectiveness of centric selection in many visual tasks has been demonstrated by lots of literature \cite{li2013learning,judd2009learning}.

\noindent \emph{- RL:} Relevant Loss \cite{rohrbach2016grounding}. We implement the model proposed by \cite{rohrbach2016grounding} as a baseline. We replace region proposals by NFoV proposal and follow the same experimental setting for fair comparison. 

\subsection{Ablation studies}

We also evaluate several variants of the proposed model. Note that \emph{w/} denotes “with”. The variants of models are listed in Tab. \ref{ablation} and the details are as follows:

\noindent \emph{- w/ Relevant Loss:} Train the model with the relevant loss proposed in \cite{rohrbach2016grounding}.

\noindent \emph{- w/ Embedded NFoV generator:} Embed the NFoV generator into feature space (See Sec. Visual Grounding Model).

\noindent \emph{- w/ Irrelevant Loss:} Train the model with the irrelevant loss (See Eq. (\ref{neg_loss})).

\noindent \emph{- w/ Data augmentation:} Train the model with augmented training data.

\begin{table}
\caption{\small Average Recall and Precision in Indoor subset. Our model with data augmentation, pretrained decoder, relevant/irrelevant loss and NFoV generator embedding achieves the highest recall/precision over all baselines and proposed methods.}
\begin{center}
\small
\resizebox{0.35\textwidth}{!}{
\begin{tabular}{ |c|c|c| } 
 \hline
 Model &  avg. Recall (\%) & avg. Precision (\%) \\ \hline\hline
 RS & 8.8 & 21.5 \\ \hline
 CS & 10.3 & 29.5 \\ \hline
 RL & 6.1 & 12.6 \\ \hline
 \hline
 RL-f & 8.3 & 19.2 \\ \hline
 D$^{\dagger}$-RL-f & 12.8 & 27.8 \\ \hline
 D$^{\dagger}$-RIL-f & 13.4 & 30.7 \\ \hline\hline
 Oracle & 51.1 & 70.0 \\ \hline
\end{tabular}
}
\end{center}
\label{performance-ablation}
\end{table}

\subsection{Evaluation Metrics}
In this work, we are interested in predicting the objects or places mentioned in subtitles. Because the objects or places in the proposed dataset are annotated by bounding boxes, they typically are not contained in an NFoV, even appear in multiple NFoVs. In this case, we propose to utilize recall and precision measurement calculated at the pixel level to evaluate the performance of our model and baselines. The recall and precision can be defined as:
\begin{equation}
Recall=\frac{\mathbf{GT_{bbox}} \cap y}{\mathbf{GT_{bbox}}}; Precision=\frac{\mathbf{GT_{bbox}} \cap y}{y},
\end{equation}
where $y$ denotes the predicted NFoV (See Eq. (\ref{predict})), $\mathbf{GT_{bbox}}$ denotes the grounding bounding boxes annotated by human annotator, and $\mathbf{GT_{bbox}} \cap y$ means compute the overlap between annotated bounding boxes and  predicted NFoV by pixels. We compute $Recall$ and $Precision$ each frame and average them across all frames in a video to acquire $avg. Recall$ and $avg. Precision$. Besides, to quantify the speed of our proposed model, we also measure $fps$ in the testing set. We conduct all experiments on a single computer with Intel(R) Core(TM) i7-5820K CPU @ 3.30GHz, 64GB RAM DDR3, and an NVIDIA TitanX GPU. 

\subsection{Results and Discussion}
\subsubsection{Ablation studies.}
The results are shown in Tab. \ref{performance-ablation} verify that our proposed method can better ground language in the panoramic image than typical visual grounding method. Moreover, the ablation studies demonstrate that all of our proposed techniques improve the performance. Our best model (D$^{\dagger}$-RIL-f) outperforms the strongest baseline (CS) by $30\%$ gain on $avg. Recall$.

\begin{figure*}[!ht]
  \centering
   \includegraphics[width=.88\textwidth]{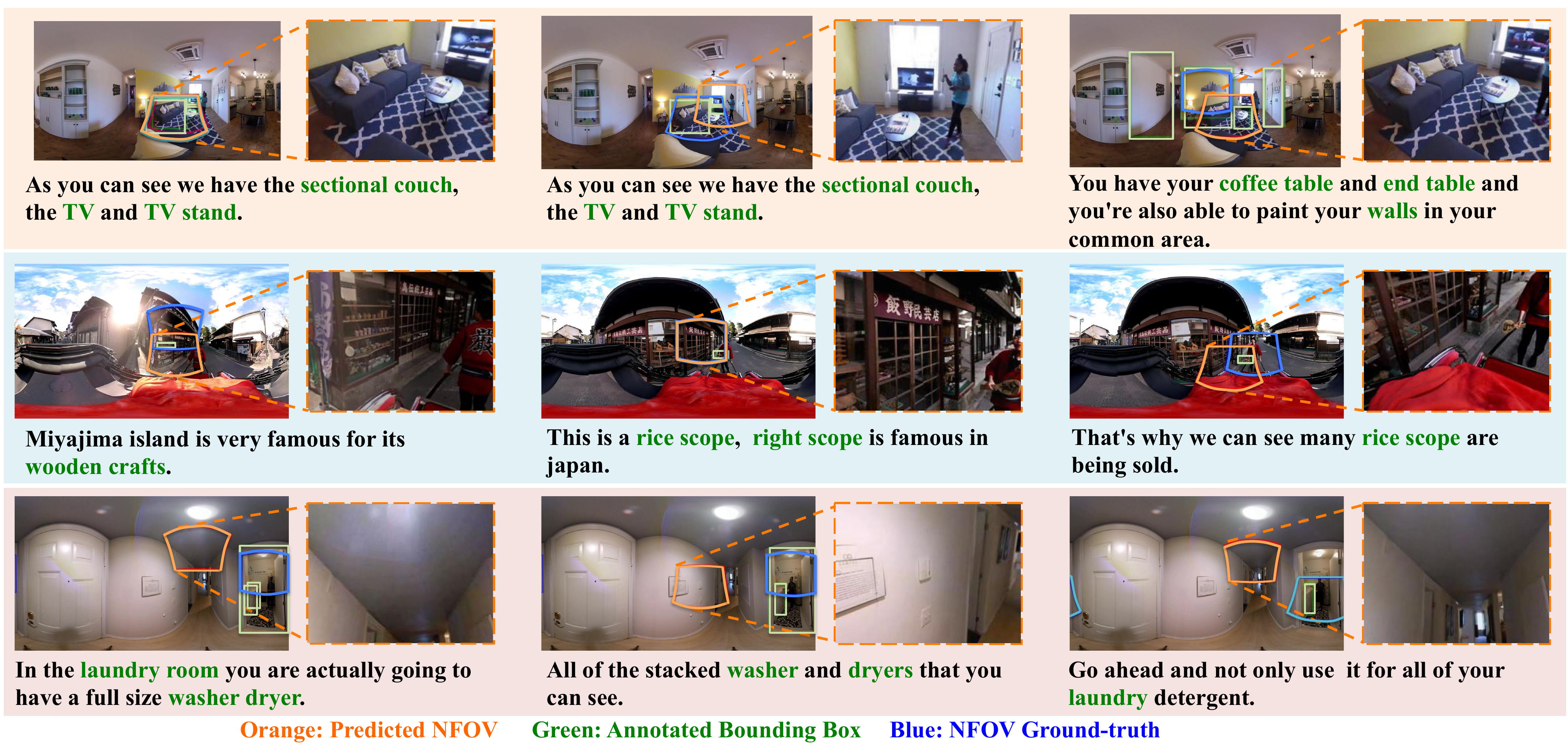}
    \caption{\small Qualitative Results on our narrated 360$^{\circ}$ videos dataset.  Top: indoor results. Middle: outdoor results. Bottom: failure case.}\label{qualitative}
\end{figure*}

\subsubsection{Model Robustness.}
Tab. \ref{performance-robust} shows that our best model (D$^{\dagger}$-RIL-f) achieves the state-of-the-art performance ($18.1\%$ and $34.6\%$ on $avg. Recall$ and $avg. Precision$) on the first narrated $360^{\circ}$ video dataset. Our best model (D$^{\dagger}$-RIL-f) is significantly better than the strongest baseline (CS) by $8.4 \%$ and $6.4 \%$ on $avg. Recall$ and $avg. Precision$, respectively. It manifests the robustness of our proposed methods.
\begin{table}
\caption{\small Average Recall and Precision on the total dataset. We test our model in total dataset (indoor + outdoor) and our model outperforms the baselines on both $avg.Recall$ and $avg.Precision$.}
\small
\begin{center}
\resizebox{0.35\textwidth}{!}{
\begin{tabular}{ |c|c|c| } 
 \hline
 Model &  avg. Recall (\%) & avg. Precision (\%) \\ \hline\hline
 RS & 8.7 & 20.6 \\ \hline
 CS & 9.7 & 28.2 \\ \hline
 RL & 5 & 12.9 \\ \hline\hline
 D$^{\dagger}$-RIL-f & 18.1 & 34.6 \\ \hline\hline
 Oracle & 51.7 & 69.4 \\ \hline
\end{tabular}
}
\end{center}
\label{performance-robust}
\end{table}

\subsubsection{Model Efficiency.}
The results are shown in Tab. \ref{performance-time} illustrate that our model is more effective and efficient than baseline. The baseline needs too many memories to train and evaluate on a single computer at the full size and $1/4$ ratio setting. Besides, our best model (D$^{\dagger}$-RIL-f) achieves 0.38 $fps$ on the full size panoramic image ($720 \times 1280$). Also, our model outperforms the baseline by a significant margin ($10$ times) at $1/16$ image size, even achieves the better performance.
\begin{table}
\caption{\small We record the testing time over total dataset and compare our proposed model with the strongest baseline. We resize the image to the different ratio and report their fps and grounding performance. The full size denotes $720 \times 1280$ panoramic image. Evidently, our proposed model is more efficient than the strongest baseline. In $1/16$ image size, ours fps is $10$ times faster than baseline but has the same performance.}
\small
\begin{center}
\resizebox{0.4\textwidth}{!}{
\begin{tabular}{ |c|c|c| } 
 \hline
 RL & fps & avg. Recall / avg. Precision (\%) \\ \hline\hline
 Full size & infeasible & infeasible \\ \hline
 $1 / 4$ & infeasible & infeasible \\ \hline
 $1 / 16$ & 0.14 & 5/12.9 \\ \hline\hline
 D$^{\dagger}$-RIL-f & fps & avg. Recall / avg. Precision (\%) \\ \hline
 Full size & 0.38 & 18.1 / 34.6 \\ \hline
 $1 / 4$ & 1.11 & 6.3 / 20 \\ \hline
 $1 / 16$ & 1.4 & 5.4/15.9 \\ \hline
\end{tabular}
}
\end{center}
\label{performance-time}
\end{table}

\subsubsection{Qualitative Result.}
To further understand the behavior of the learned model, we show the qualitative results in Fig. \ref{qualitative}. By observing this figure, we find that our proposed model can ground the phrase in the corresponding subtitle. In the indoor set, there may have multiple references in one subtitle. It is shown in Fig. \ref{qualitative} that the predicted viewpoint is successfully find the coach, TV and table.

\section{Conclusion}
We introduce a new NFoV grounding task which aims to ground subtitles (i.e., natural language phrases) into corresponding views in each $360^{\circ}$ video. To tackle this task, we propose a novel model with two main innovations: (1) train the network with both of relevant and irrelevant soft-attention visual features, and (2) apply NFoV proposal in feature space for saving time-consuming, memory usage, and making end-to-end training feasible. We achieve the best performance on both recall and precision measurement. In the future, we plan to extend the dataset and model to joint story-telling and NFoV grounding in $360^{\circ}$ touring videos.

\section{Acknowledgments}
We thank MOST 106-2221-E-007-107, MOST 106-3114-E-007-008, Microsoft Research Asia and MediaTek for their supports. We thank Hsien-Tzu Cheng and Tseng-Hung Chen for
helpful comments and discussion.

\bibliographystyle{aaai}


\end{document}